\documentclass[11pt]{article}
\usepackage{graphicx}
\usepackage{color}
\usepackage{amsmath}
\usepackage{amssymb}
\usepackage{amscd}
\usepackage{bbm}
\newcommand{\R}{\mathbb{R}}
\newcommand{\inr}[1]{\bigl< #1 \bigr>}

\newcommand{\E}{\mathbb{E}}

\newcommand{\eps}{\varepsilon}

\newtheorem{Theorem}{Theorem}[section]
\newtheorem{Lemma}[Theorem]{Lemma}

\newtheorem{Definition}[Theorem]{Definition}

\newtheorem{Corollary}[Theorem]{Corollary}
\newtheorem{Remark}[Theorem]{Remark}

\newtheorem{Question}[Theorem]{Question}
\numberwithin{equation}{section}

\def \proof {\noindent {\bf Proof.}\ \ }

\def \endproof
{{\mbox{}\nolinebreak\hfill\rule{2mm}{2mm}\par\medbreak}}

\begin{document}
\title{{Column normalization of a random measurement matrix}}
\author{Shahar Mendelson${}^{1,2}$}

\footnotetext[1]{Department of Mathematics, Technion, I.I.T., Haifa, Israel and Mathematical Sciences Institute, The Australian National University, Canberra, Australia, Email:
shahar@tx.technion.ac.il}
\footnotetext[2]{Supported in part by the Israel Science Foundation.}

\maketitle

\begin{abstract}
In this note we answer a question of G. Lecu\'{e}, by showing that column normalization of a random matrix with iid entries need not lead to good sparse recovery properties, even if the generating random variable has a reasonable moment growth. Specifically, for every $2 \leq p \leq c_1\log d$ we construct a random vector $X \in \R^d$ with iid, mean-zero, variance $1$ coordinates, that satisfies $\sup_{t \in S^{d-1}} \|\inr{X,t}\|_{L_q} \leq c_2\sqrt{q}$ for every $2\leq q \leq p$.
 We show that if $m \leq c_3\sqrt{p}d^{1/p}$ and $\tilde{\Gamma}:\R^d \to \R^m$ is the column-normalized matrix generated by $m$ independent copies of $X$, then with probability at least $1-2\exp(-c_4m)$, $\tilde{\Gamma}$ does not satisfy the exact reconstruction property of order $2$.
\end{abstract}

\section{Introduction}
\emph{Sparse Recovery} is one of the most important research topics in modern signal processing. It focuses on the possibility of identifying a sparse signal---i.e., a signal that is supported on relatively few coordinates in $\R^d$ relative to the standard basis---using linear measurements. We refer the reader to the books \cite{MR2807761,fora13} for extensive surveys on sparse recovery and related topics.

In a basic sparse recovery problem one pre-selects an $m \times d$ matrix $\Gamma$ that generates the given data. For an unknown (sparse) vector $v$, the coordinates of the vector $\Gamma v$ are the $m$ linear measurements of $v$ one may use for recovery. The hope is that for a well chosen $\Gamma$, the resulting $m$ linear measurements would be enough to identify $v$, and because $v$ is sparse, the number of measurements required for recovery should be significantly smaller than the dimension $d$.

One of the main achievements of the theory of sparse recovery was the introduction of a convex optimization problem called \emph{basis pursuit}, which is an effective recovery procedure: it selects $t \in \R^d$ that solves the minimization problem
\begin{equation} \label{eq:BP}
\min\|t\|_1 \ \ \ {\rm subject \ to} \ \ \ \Gamma v = \Gamma t,
\end{equation}
where we denote by $\|x\|_p = (\sum_{j=1}^d |x_j|^p)^{1/p}$.

Extensive effort has been devoted to the question of finding conditions on the measurement matrix $\Gamma$ that ensure the recovery of any sparse vector; more accurately, one would like to guarantee that for every $s$-sparse vector $v$, the $\ell_1$ minimization problem \eqref{eq:BP} has a unique solution---$v$ itself.
\begin{Definition} \label{def:ERP}
Let $\Sigma_s$ be the set of $s$-sparse vectors in $\R^d$. A matrix $\Gamma:\R^d \to \R^m$ satisfies the exact reconstruction property of order $s$ if for every $v \in \Sigma_s$ there is a unique solution to the minimization problem \eqref{eq:BP} and that unique solution is $v$.
\end{Definition}
Because measurements are `expensive', one would like to find matrices $\Gamma$ that satisfy the exact recovery property of order $s$ with the smallest number of measurements (rows) possible. One may show that if $\Gamma$ satisfies the exact reconstruction property of order $s$, then it must have at least $m \sim s\log(ed/s)$ rows. Moreover, typical realizations of various random matrices with $\sim s \log(ed/s)$ rows indeed satisfy the exact reconstruction property of order $s$ (see, e.g., \cite{fora13}). Therefore, the optimal number of measurements required for the exact reconstruction property of order $s$ is $m \sim s\log(ed/s)$, and that number serves as the benchmark for an optimal measurement matrix.

\vskip0.4cm
The question we are interested in has to do with the normalization of the columns of the measurement matrix. It is often assumed in literature that the columns of $\Gamma$ have unit Euclidean norm (see, for example, \cite{MR2807761} and \cite{fora13} and references therein); i.e., if $\{e_1,...,e_d\}$ is the standard basis in $\R^d$ then $\|\Gamma e_j\|_2 =1$ for $1 \leq j \leq d$. Column normalization appears frequently in various notions used in the study of the exact reconstruction property. Among these well-studied notions are \emph{coherence} \cite{fora13}; the \emph{restricted eigenvalues condition} \cite{MR2533469}; and the \emph{compatibility condition} \cite{MR2807761}. Moreover, in many real-world applications, measurement matrices with normalized columns tends to perform better than matrices whose columns have not been normalized.

While column normalization seems a natural idea, it adds substantial technical difficulties when studying random measurement matrices: normalizing the columns of a matrix with independent rows introduces additional dependencies. Despite the added difficulties, the results of \cite{LM-JEMS} highlight the possibility that column normalization may still have a significant role to play in the context of random measurement matrices, particularly in heavy-tailed situations.

To formulate the results of \cite{LM-JEMS} and explain their connection to column-normalization we need the following definition:
\begin{Definition}
Let $x$ be a random variable. Given an integer $m \leq d$, let $(x_{ij})$, $1 \leq i \leq m$, $1 \leq j \leq d$ be $md$ independent copies of $x$. The random matrix generated by $x$ is $\Gamma=(x_{ij})_{1 \leq i \leq m, 1 \leq j \leq d}$. Also, we denote by $X=(x_j)_{j=1}^d$ a vector with $d$ independent copies of $x$; thus the rows of $\Gamma$ are $m$ independent copies of $X$.
\end{Definition}
The following result from \cite{LM-JEMS} is a construction of random matrices generated by seemingly nice random variables, but despite that the matrices exhibit poor reconstruction properties.
\begin{Theorem} \label{thm:LM-JEMS2} 
There exist absolute constants $c_1,c_2$ and $c_3$ for which the following holds. For every $2 < p \leq c_1 \log d$ there is a mean-zero, variance one random variable $x$ that satisfies
\begin{description}
\item{$\bullet$} For every $2 \leq q \leq p$ and every $t \in S^{d-1}$,
$$
\|\inr{X,t}\|_{L_q} \leq c_2 \sqrt{q} \|\inr{X,t}\|_{L_2} = c_2 \sqrt{q}.
$$
\item{$\bullet$} If $m \leq c_3\sqrt{p}(d/\log d)^{1/p}$ then with probability $1/2$, $\Gamma$ does not satisfy the exact reconstruction property of order $1$.
\end{description}
\end{Theorem}

Theorem \ref{thm:LM-JEMS2} implies that without assuming that each $\inr{X,t}$ has a subgaussian moment growth\footnote{Recall that a characterization of an $L$-subgaussian random variable is that $\|z\|_{L_p} \leq L\sqrt{p}\|z\|_{L_2}$ for every $p \geq 2$} up to the $p$-moment for $p$ close to $\log d$, the resulting measurement matrix is suboptimal. Indeed, under a modest assumption, say that $\|\inr{X,t}\|_{L_4} \leq c\|\inr{X,t}\|_{L_2}$ for every $t \in \R^d$, the recovery of $1$-sparse vectors requires at least $(d/\log d)^{1/4}$ measurements. And, if $p = (\log d)/(\beta \log \log d)$ for $\beta$ large enough, then the number of measurements required for the recovery of $1$-sparse vectors is at least $\sim \log^{c\beta}d$, which  is suboptimal when $c\beta>1$.

To put Theorem \ref{thm:LM-JEMS2} in some perspective, it is complemented by a positive result, once linear forms have enough subgaussian moments. 
\begin{Theorem} \label{thm:LM-JEMS1}
Let $x$ be a mean-zero, variance one random variable. Assume that for every $2 \leq q \leq c_4\log d$ and every $t \in S^{d-1}$,
\begin{equation} \label{eq:subgaussian-growth}
\|\inr{X,t}\|_{L_q} \leq L \sqrt{q} \|\inr{X,t}\|_{L_2} = L \sqrt{q}.
\end{equation}
If
$$
m \geq c_5 s\log(ed/s),
$$
then with probability at least $1-1/d^{c_6}-2\exp(-c_7m)$, $\Gamma$ satisfies the exact reconstruction property of order $s$. Here, $c_4$ in an absolute constant and $c_5,c_6$ and $c_7$ are constants that depend only on $L$.
\end{Theorem}
It follows from Theorem \ref{thm:LM-JEMS1} that if $X$ has a slightly better moment growth condition than in Theorem \ref{thm:LM-JEMS2}---a subgaussian growth up to $p \sim \log d$---the random measurement matrix generated by $x$ satisfies the exact reconstruction property of order $s$, for the optimal number of measurements $m \sim s\log(ed/s)$.

\vskip0.4cm
The connection with column-normalization arises from the main observation used in the proof of Theorem \ref{thm:LM-JEMS1}:
\begin{Lemma} \label{lemma:upper-bound}
Recall that $\Sigma_s$ denotes the set of $s$-sparse vectors in $\R^d$. Let $\Gamma : \R^d \to \R^m$. If
\begin{description}
\item{(a)} $\|\Gamma x\|_2 \geq \alpha \|x\|_2$ for every $x \in \Sigma_s$,
\item{(b)} $\|\Gamma e_j\|_2 \leq \beta$ for every $j \in \{1,...,d\}$,
\end{description}
and $s_1=\lfloor \alpha^2(s-1)/(4\beta^2)\rfloor-1$, then $\Gamma$ satisfies the exact reconstruction property of order $s_1$.
\end{Lemma}

Lemma \ref{lemma:upper-bound} gives a clear motivation for considering column-normalized random measurement matrices, and that motivation grows stronger when taking into account the proof of Theorem \ref{thm:LM-JEMS1}. It turns out that the `bottleneck' in the proof is the upper bound on $\max_{1 \leq j \leq d} \|\Gamma e_j\|_2$, while guaranteeing (a) requires a rather minimal \emph{small-ball} assumption. Therefore, the seemingly more restrictive condition (a) is almost universally true (see \cite{MenACM,LM-JEMS} for more details) and (b) is the only place in which the moment growth assumption is used in the proof of Theorem \ref{thm:LM-JEMS1}.

Clearly, column normalization resolves the issue of an upper estimate on $\max_{1 \leq j \leq d} \|\Gamma e_j\|_2$. That, and the fact that (a) is true under minimal assumptions has led G. Lecu\'{e} \cite{Lec} to ask whether with column normalization, the moment growth condition \eqref{eq:subgaussian-growth} can be relaxed significantly, leading to a much stronger version of Theorem \ref{thm:LM-JEMS1}.
\begin{Question} \label{Qu:Guillaume}
Let $x$ be a mean-zero, variance $1$ random variable, set $\Gamma$ to be the $m \times d$ matrix generated by $x$ and let $\tilde{\Gamma}$ be the column-normalized matrix generated by $x$. Thus, the entries of $\tilde{\Gamma}$ are
$$
\tilde{\Gamma}_{ij} = \frac{x_{ij}}{\left(\sum_{\ell=1}^m x_{\ell j}^2 \right)^{1/2}} = \frac{\Gamma_{ij}}{\|\Gamma e_j\|_2}.
$$
If $\|\inr{X,t}\|_{L_4} \leq L\|\inr{X,t}\|_{L_2}$ for every $t \in \R^d$ and $m = c(L)s\log(ed/s)$, does $\tilde{\Gamma}$ satisfy the exact reconstruction property of order $s$ with high probability?
\end{Question}

Our main result is a version of Theorem \ref{thm:LM-JEMS2} for a column-normalized matrix generated by well chosen random variable, showing that the answer to question \ref{Qu:Guillaume} is negative.

\begin{Theorem} \label{thm:main}
There exist absolute constants $c_1,c_2$ and $c_3$ for which the following holds.
For every $2 \leq p \leq \log d$ there is a symmetric, variance $1$ random variable $x$ with the following properties:
\begin{description}
\item{$\bullet$} If $x_1,...,x_d$ are independent copies of $x$ and $X=(x_j)_{j=1}^d$, then for every $t \in S^{d-1}$ and  every $2 \leq q \leq p$, $\|\inr{X,t}\|_{L_q} \leq c_1\sqrt{q}\|\inr{X,t}\|_{L_2}$.
\item{$\bullet$} If $m \leq c_2\sqrt{p}d^{1/p}$, then with probability at least $1-2\exp(-c_3m)$, the $m \times d$  column-normalized matrix generated by $x$ does not satisfy the exact reconstruction property of order $2$.
\end{description}
\end{Theorem}

Theorem \ref{thm:main} answers Question \ref{Qu:Guillaume} in the negative: column normalization does not improve the poor behaviour described in Theorem \ref{thm:LM-JEMS2}. Indeed, for $p=4$, linear forms $\inr{X,t}$ satisfy an $L_2-L_4$ norm equivalence, but the recovery of $2$-sparse vectors using $\tilde{\Gamma}$ requires at least $m \sim d^{1/4}$ measurements --- significantly larger than the optimal number of measurements, $m \sim \log d$. Moreover, if $\beta>1$ and $p = (\log d)/{\beta \log \log d}$, then although $\|\inr{X,t}\|_{L_q} \lesssim \sqrt{q}\|\inr{X,t}\|_{L_2}$ for every $2 \leq q \leq p$, the recovery of $2$-sparse vectors using $\tilde{\Gamma}$ requires at least $m \sim \log^{c\beta} d$ measurements, which, again, is suboptimal when $c\beta>1$.

\begin{Remark}
Theorem \ref{thm:main} actually improves the estimates from Theorem \ref{thm:LM-JEMS2}: a logarithmic factor in the bound on the number of measurements is removed, and the probability estimate is significantly better: $1-2\exp(-cm)$ rather than constant probability. 
\end{Remark}

Let us mention the straightforward observation that a version of Theorem \ref{thm:LM-JEMS1} holds for column-normalized matrices as well.
\begin{Theorem} \label{thm:upper-normalized}
Let $x$ be and $L$ be as in Theorem \ref{thm:LM-JEMS1} and let $\tilde{\Gamma}$ be the column-normalized measurement matrix generated by $x$.
If $m \geq c_8(L) s\log(ed/s)$, then with probability at least $1-1/d^{c_9(L)}-2\exp(-c_{10}(L)m)$, $\tilde{\Gamma}$ satisfies the exact reconstruction property of order $s$. 
\end{Theorem}
Theorem \ref{thm:upper-normalized} is an immediate consequence of the proof of Theorem \ref{thm:LM-JEMS1}; its proof is presented in Appendix \ref{sec:proof-upper} merely for the sake of completeness.

\section{Proof of Theorem \ref{thm:main}}
Let $\eps$ be a symmetric, $\{-1,1\}$-valued random variable, set $\eta$ to be a $\{0,1\}$-valued random variable with mean $\delta$ and let $R>0$; the values of $\delta$ and $R$ will be specified later. Let
$$
x=\eps \cdot \max\{1,\eta R\},
$$
let $x_1,...,x_d$ be independent copies of $x$ and set $X=(x_1,...,x_d)$.

Let us identify conditions under which $X$ satisfies the first part of Theorem \ref{thm:main}.

\begin{Lemma} \label{lemma:single-X}
There exists an absolute constant $c_0$ for which the following holds.
Assume that $\delta<1/2$ and that there is $L \geq 1$ such that for every $2 \leq q \leq p$, $R\delta^{1/q} \leq L \sqrt{q}$. Then for every $t \in \R^d$ and every $2 \leq q \leq p$,
$$
\|\inr{X,t}\|_{L_q} \leq c_0L\sqrt{q}\|\inr{X,t}\|_{L_2}.
$$
Moreover, for every $t \in \R^d$, $\|\inr{X,t}\|_{L_2} =c_1\|t\|_2$, and $1/\sqrt{2} \leq c_1 \leq 2L$.

In particular, $X/c_1$ is an isotropic random vector and for every $t \in \R^d$, $\inr{X,t}$ exhibits a $c_0 L$-subgaussian moment growth up to the $p$-th moment.
\end{Lemma}

The proof of Lemma \ref{lemma:single-X} is based on a simple comparison argument:
\begin{Lemma} \label{lemma:dominating-psi-2}
Let $x_1,...,x_d$ be centred, independent random variables and assume $z_1,...,z_d$ are also centred and independent. If $p$ is even and for every $1 \leq j \leq d$ and every $1 \leq q \leq p$, $\|x_i\|_{L_q} \leq L\|z_i\|_{L_q}$, then for every $t \in \R^d$,
$$
\|\sum_{j=1}^d t_j x_j\|_{L_p} \leq L \|\sum_{t=1}^d t_j z_j\|_{L_p}.
$$
\end{Lemma}
\proof By a standard  symmetrization argument we may also assume that $z_1,...,z_d$ and $x_1,...,x_d$ are symmetric. Therefore,
$$
\E (\sum_{j=1}^d t_j x_j)^p = \E \sum_{\vec{\beta}}  c_{\vec{\beta}} \prod_{j=1}^d t_j^{\beta_j} x_j^{\beta_j} = \sum_{\vec{\beta}} c_{\vec{\beta}} \prod_{j=1}^d t_j^{\beta_j} \E x_j^{\beta_j},
$$
with the sum taken over all choices of $\vec{\beta}=(\beta_1,...,\beta_d) \in \{0,...,p\}^d$, where $\sum_{j=1}^d \beta_j =p$ and $c_{\vec{\beta}}$ is the appropriate multinomial coefficient. Since $x_1,...,x_d$ are symmetric, the only products that do not vanish are when $\beta_1,...,\beta_d$ are even, and if $\beta_1,...,\beta_d$ are even then
$$
\prod_{j=1}^d t_j^{\beta_j} \E x_j^{\beta_j} \leq \prod_{j=1}^d t_j^{\beta_j} L^{\beta_j} \E z_j^{\beta_j}.
$$
Therefore,
$$
\sum_{\vec{\beta}} c_{\vec{\beta}} \prod_{j=1}^d t_j^{\beta_j} \E x_j^{\beta_j} \leq L^p \sum_{\vec{\beta}} c_{\vec{\beta}} \prod_{j=1}^d t_j^{\beta_j} \E z_j^{\beta_j} = L^p \E(\sum_{j=1}^d t_j z_j)^p.
$$
\endproof

\noindent {\bf Proof of Lemma \ref{lemma:single-X}.} Observe that $x=\eps\max\{1,R\eta\}$ is mean-zero and that $\E x^2 = 1 \cdot (1-\delta) + R^2 \delta$. Hence, if $\delta \leq 1/2$ and $R^2 \delta \leq 2L^2$ then $1/2 \leq \E x^2 \leq 4L^2$---and the ``moreover" part of the claim follows.

Turning to the first part of the claim, let $x_1,...,x_d$ be independent copies of $x$, set $g$ to be a standard gaussian random variable and let $g_1,...,g_d$ be independent copies of $g$. Recall that for every $2 \leq q \leq p$, $R\delta^{1/q} \leq L \sqrt{q}$, and observe that
$$
(\E|x|^q)^{1/q} \leq  1+R \delta^{1/q} \leq 2L\sqrt{q} \leq c_1L(\E|g|^q)^{1/q}.
$$
Therefore, $(x_1,...,x_d)$ and $(g_1,...,g_d)$ satisfy the conditions of Lemma \ref{lemma:dominating-psi-2} with a constant $c_1L$. Applying Lemma \ref{lemma:dominating-psi-2}, it follows that for every $t \in S^{d-1}$ and every $2 \leq q \leq p$,
$$
\|\sum_{j=1}^d t_j x_j \|_{L_q} \leq c_1L \||\sum_{j=1}^d t_j g_j \|_{L_q} \leq c_2L\sqrt{q};
$$
thus, $\|\inr{X,t}\|_{L_q} \leq c_3L\sqrt{q} \|\inr{X,t}\|_{L_2}$.
\endproof

\vskip0.4cm
The key part in the construction is the following lemma which describes the typical structure of the matrix generated by $x$,
$$
\Gamma=(x_{ij})_{1 \leq i \leq m, 1 \leq j \leq d}:\R^d \to \R^m.
$$
\begin{Lemma} \label{lemma:main}
There exist absolute constants $c_1,c_2,c_3$ and $c_4$ for which the following holds. Let $\delta = c_1/d$ and $R \geq c_2m$.
Then, with probability at least $1-2\exp(-c_3m)$:
\begin{description}
\item{(1)} there are indices $j_1 \not= j_2 \in \{1,...,d\}$ and $1 \leq \ell \leq m$ such that $\eta_{\ell j_1}=\eta_{\ell j_2}=1$ and for $i \not= \ell$, $\eta_{\ell j_1}=\eta_{\ell j_2}=0$;
\item{$(2)$} there is a subset $J \subset \{1,...,d\}$ of cardinality  $|J|=2m$ such that $\eta_{ij}=0$ for every $j \in J$ and $1 \leq i \leq m$;
\item{$(3)$} we have that $c_4 B_2^m \subset \Gamma B_1^J$, where $B_1^J=\{x=\sum_{j \in J} x_je_j : \|x\|_1 \leq 1\}$ and $B_2^m = \{x \in \R^m : \|x\|_2 \leq 1\}$.
\end{description}
\end{Lemma}

\begin{Corollary} \label{cor:not-ERP}
If $\Gamma$ satisfies Lemma \ref{lemma:main} then its column-normalized version $\tilde{\Gamma}$ does not satisfy the exact reconstruction property of order $2$.
\end{Corollary}

\proof Using the notation of Lemma \ref{lemma:main} and by its first part, $\|\Gamma e_{j_1}\|_2=\|\Gamma e_{j_2}\|_2 = (R^2+m-1)^{1/2}$; hence, if we denote by $\{f_1,...,f_m\}$ the standard basis of $\R^m$,
$$
\tilde{\Gamma}e_{j_1} = \frac{1}{(R^2+m-1)^{1/2}} \bigl(\eps_{\ell j_1} R f_\ell+ \sum_{i \not=\ell} \eps_{ij_1} f_i \bigr)
$$
and
$$
\tilde{\Gamma} e_{j_2} = \frac{1}{(R^2+m-1)^{1/2}} \bigl(\eps_{\ell j_2} R f_\ell + \sum_{i \not= \ell} \eps_{ij_2} f_i \bigr).
$$
If $\eps_{\ell j_1} \not=\eps_{\ell j_2}$ set $v=(e_{j_1}+e_{j_2})/2$; otherwise, set $v=(e_{j_1}-e_{j_2})/2$. In either case, $v$ is $2$-sparse. Let $w=\tilde{\Gamma} v$ and observe that the coordinates of $w$ satisfy that
$$
w_\ell = 0 \ \ \ {\rm and} \ \ \ w_i^2 \leq \frac{1}{R^2+m-1}  \ \ \ {\rm for} \ \ i \not = \ell;
$$
therefore,
$$
\tilde{\Gamma}v  \in \frac{\sqrt{m}}{R} B_2^m.
$$
Next, let $J$ be the set of coordinates given by the second part of Lemma \ref{lemma:main}. Clearly, $j_1,j_2 \not\in J$ and
$$
\Gamma^J = (x_{ij})_{1 \leq i \leq m, j \in J}=(\eps_{ij})_{1 \leq i \leq m, j \in J}
$$
is an $m \times 2m$ Bernoulli matrix. Therefore,
$$
\tilde{\Gamma}^J=(\tilde{\Gamma}_{ij})_{1 \leq i \leq m, j \in J} = \frac{\Gamma^J}{\sqrt{m}}.
$$
Observe that $\tilde{\Gamma} B_1^J=\tilde{\Gamma}^J B_1^J$ and by the third part of Lemma \ref{lemma:main}
$$
\frac{c}{\sqrt{m}}B_2^m \subset \frac{1}{\sqrt{m}}\Gamma B_1^J
$$
for an absolute constant $c$. 

Hence, if $\sqrt{m}/R \leq c/\sqrt{m}$ then $\tilde{\Gamma}v \in \tilde{\Gamma}B_1^J$.
Since $\|v\|_1=1$ and $v \not \in B_1^J$, it is evident that $v$ is not the unique solution of the minimization problem
$$
\min \|t\|_1 \ \ \  {\rm subject \ to \ } \ \ \ \tilde{\Gamma}v=\tilde{\Gamma}t
$$
and $\tilde{\Gamma}$ does not satisfy the exact reconstruction property of order $2$.
\endproof

The proof of Lemma \ref{lemma:main} uses a standard fact on iid $\{0,1\}$-valued random variables: if $W_1,...,W_d$ are independent copies of a $\{0,1\}$-valued random variable $W$ and $\E W = \mu$ then with probability at least $1-2\exp(-c\mu d)$, $\mu d /2 \leq |\{j : W_j =1\}| \leq 3\mu d/2$.

\vskip0.4cm

\noindent{\bf Proof of Lemma \ref{lemma:main}.} Let $\eta_1,...,\eta_m$ be independent copies of $\eta$, let $Y$ be the indicator of the event
$$
\exists \ell \in \{1,...,m\} \ \ \ \ \eta_\ell=1 \ \ \ {\rm and} \ \ \ \eta_i =0 \ {\rm for \ every \ } i \not=\ell.
$$
Observe that $\E Y =m\delta(1-\delta)^{m-1}$ and that if $Y_1,...,Y_d$ are independent copies of $Y$ and $\E Y \geq 2m/d$ then with probability at least $1-2\exp(-c_1m)$, $|\{i : Y_i = 1\}| > m$. In particular, on that event, the matrix $(\eta_{ij})_{1 \leq i \leq m, 1 \leq j \leq d}$ has at least two identical columns, each with a single entry of $1$.  Therefore, the first part of Lemma \ref{lemma:main} holds if
\begin{equation} \label{eq:cond-1}
m\delta(1-\delta)^{m-1} \geq \frac{2m}{d}
\end{equation}

For the second part of the lemma, let $Z$ be the indictor of the event
$$
\eta_i =0 \ \ \ {\rm for \ every \ }  1 \leq i \leq m
$$
and note that $\E Z = (1-\delta)^m$. If $Z_1,...,Z_d$ are independent copies of $Z$ and $\E Z \geq 4 m/d$ then with probability at least $1-2\exp(-c_2 m)$, $|\{i: Z_i=1\}| \geq 2m$. Hence, if
\begin{equation} \label{eq:cond-2}
(1-\delta)^m \geq \frac{4m}{d},
\end{equation}
then with probability at least $1-2\exp(-c_2m)$, there is $J \subset \{1,...,d\}$ and for every $j \in J$ and every $1 \leq i \leq m$, $\eta_{ij}=0$.

Turning to the third part of the lemma, and by applying the second part, we have that for $(i,j) \in \{1,...,m\} \times J$, $x_{ij}=\eps_{ij}$. Let $\Gamma^J=(\eps_{ij})_{1 \leq i \leq m, j \in J}$ and recall that $(\eps_{ij})$ are independent of $(\eta_{ij})$. Therefore, by Corollary 4.1 from \cite{LPRP}, there are absolute constants $c_3$ and $c_4$ for which, with probability at least $1-2\exp(-c_3m)$,
$$
c_4 B_2^m \subset \Gamma B_1^J.
$$

Finally, all that remains is to see when \eqref{eq:cond-1} and \eqref{eq:cond-2} are satisfied. It is straightforward to verify that if $\delta = 2\theta/d$ for $2 \leq \theta \leq c_5\frac{d}{m} \log(ed/m)$ then \eqref{eq:cond-1} holds, and if $\delta \leq \frac{c_6}{m} \log(ed/m)$ then \eqref{eq:cond-2} holds. Therefore, both conditions are satisfied with the choice of $\delta=c/d$ for a suitable absolute constant $c>1$, as long as $m \lesssim d$.
\endproof

To complete the proof of Theorem \ref{thm:main}, let $\delta=c_1/d$ as above, set $p > 2$ and put $R=\sqrt{p}d^{1/p}$---a choice which complies with the conditions of Lemma \ref{lemma:main} as long as
\begin{equation} \label{eq:fin-cond-1}
m \leq c_2 \sqrt{p} d^{1/p}.
\end{equation}
It follows from Corollary \ref{cor:not-ERP} that with probability at least $1-2\exp(-c_3m)$, the column-normalized matrix $\tilde{\Gamma}$ generated by $x=\eps \max\{1,R\eta\}$ does not satisfy the exact reconstruction property of order $2$. To complete that proof, all that remains is to show that $x$ also satisfies the assumptions of Lemma \ref{lemma:single-X}: that $R\delta^{1/q} \leq L \sqrt{q}$
for every $2 \leq q \leq p$ and for an absolute constant $L$.

To that end, let $\phi(x)=\sqrt{x}(d/c_1)^{1/x}$ and observe that $\phi(x)$ is decreasing when $2 \leq x \leq 2\log(d/c_1)$; hence, $\phi(p)/\phi(q) \leq 1$ for every $2 \leq q \leq p$ as long as $p \leq 2\log(d/c_1)$. Therefore, if we set $L=c_1$ then $R\delta^{1/q} \leq  \sqrt{q}$ for every $q \leq p$, as required.
\endproof

\begin{footnotesize}
\bibliographystyle{plain}
\bibliography{biblio}

\begin{thebibliography}{1}

\bibitem{MR2533469}
Peter~J. Bickel, Ya'acov Ritov, and Alexandre~B. Tsybakov.
\newblock Simultaneous analysis of lasso and {D}antzig selector.
\newblock {\em Ann. Statist.}, 37(4):1705--1732, 2009.

\bibitem{MR2807761}
Peter B\"uhlmann and Sara van~de Geer.
\newblock {\em Statistics for high-dimensional data}.
\newblock Springer Series in Statistics. Springer, Heidelberg, 2011.
\newblock Methods, theory and applications.

\bibitem{fora13}
Simon Foucart and Holger Rauhut.
\newblock {\em A mathematical introduction to compressive sensing}.
\newblock Applied and Numerical Harmonic Analysis. Birkh\"auser/Springer, New
  York, 2013.

\bibitem{Lec}
Guillaume Lecu\'{e}.
\newblock Private communication.

\bibitem{LM-JEMS}
Guillaume Lecu\'{e} and Shahar Mendelson.
\newblock Sparse recovery under weak moment assumptions.
\newblock {\em J. Eur. Math. Soc. (JEMS)}, 19(3):881--904, 2017.

\bibitem{LPRP}
A.~E. Litvak, A.~Pajor, M.~Rudelson, and N.~Tomczak-Jaegermann.
\newblock Smallest singular value of random matrices and geometry of random
  polytopes.
\newblock {\em Adv. Math.}, 195(2):491--523, 2005.

\bibitem{MenACM}
Shahar Mendelson.
\newblock Learning without concentration.
\newblock {\em Journal of the ACM (JACM)}, 62(3):21, 2015.

\end{thebibliography}
\end{footnotesize}

\newpage
\appendix
\section{Proof of Theorem \ref{thm:upper-normalized}} \label{sec:proof-upper}
The proof is a direct consequence of the argument used in the proof of Theorem \ref{thm:LM-JEMS1}. Thanks to column normalization, $\tilde{\Gamma}$ satisfies (b) in Lemma \ref{lemma:upper-bound} for $\beta=1$. All that is left to verify is (a) for $\alpha$ which is a constant that depends only on $L$. 

The proof of Theorem \ref{thm:LM-JEMS1} shows that if $\Gamma$ has $m \geq c_1(L)s \log (ed/s)$ independent rows that are distributed as $X$ then with probability at least $1-2\exp(-c_2(L)m)$,
$$
\inf_{t \in \Sigma_s} \|\Gamma t\|_2^2 = \inf_{t \in \Sigma_s}  \sum_{i=1}^m \inr{X_i,t}^2 \geq c_3(L)m\|t\|_2^2.
$$
Also, with probability at least $1-1/d^{c_4(L)}$,
$$
\max_{1 \leq j \leq d} \|\Gamma e_j\|_2 \leq c_5(L)\sqrt{m}.
$$

For every $t \in \Sigma_s$, set
$$
\tilde{t} = \sum_{j=1}^d \frac{t_j}{\|\Gamma e_j\|_2} e_j,
$$
which is also an $s$-sparse vector. Observe that $\tilde{\Gamma} t = \Gamma \tilde{t}$, implying that
$$
\|\tilde{\Gamma} t\|_2^2 \geq c_3 m\sum_{j=1}^d \frac{t_j^2}{\|\Gamma e_j\|_2^2} \geq \frac{c_3}{c_5^2}\|t\|_2^2,
$$
and (a) from Lemma \ref{lemma:upper-bound} is verified for the matrix $\tilde{\Gamma}$ for $\alpha=c_6(L)$.
\endproof

\end{document}